\documentclass{article}

\usepackage[nonatbib,preprint]{neurips_2020}

\usepackage[utf8]{inputenc} 
\usepackage[T1]{fontenc}    
\usepackage{hyperref}       
\usepackage{url}            
\usepackage{booktabs}       
\usepackage{amsfonts}       
\usepackage{nicefrac}       
\usepackage{microtype}      

\usepackage{amsmath}
\usepackage{amssymb}

\usepackage{bm}
\usepackage{float}
\usepackage{txfonts}
\usepackage{subfigure}
\usepackage{caption}
\usepackage{verbatim}
\usepackage{graphicx}
\usepackage{amsmath}
\usepackage{booktabs}
\usepackage{bm}
\usepackage{float}
\usepackage{subfigure}
\usepackage{wrapfig}
\usepackage{multirow}
\usepackage{setspace}
\usepackage{color}
\definecolor{gray}{rgb}{0.35,0.35,0.35}
\definecolor{blue}{rgb}{0,0,1}
\definecolor{red}{rgb}{1,0,0}
\definecolor{orange}{rgb}{0.75, 0.4, 0}
\definecolor{purple}{rgb}{0.5, 0.0, 0.5}

\usepackage{verbatim}

\title{DO-Conv: Depthwise Over-parameterized Convolutional Layer}

\author{%
  Jinming Cao\textsuperscript{2,1}\thanks{Joint first authors with equal contributions.} \hspace{0.4in} Yangyan Li\textsuperscript{1}\footnotemark[1] \hspace{0.4in} Mingchao Sun\textsuperscript{2,1} \hspace{0.4in} Ying Chen\textsuperscript{1}
  \\
  \textbf{Dani Lischinski\textsuperscript{3} \hspace{0.25in} Daniel Cohen-Or\textsuperscript{4} \hspace{0.25in} Baoquan Chen\textsuperscript{5} \hspace{0.25in} Changhe Tu\textsuperscript{2}} 
  \AND
  \textsuperscript{1}Alibaba Group\hspace{0.9in} \textsuperscript{2}Shandong University
  \\
  \textbf{\textsuperscript{3}The Hebrew University of Jerusalem\hspace{0.15in}
  \textsuperscript{4}Tel Aviv University\hspace{0.15in}
  \textsuperscript{5}Peking University}
  \\
}

\begin{document}

\maketitle

\setcounter{footnote}{0}

\begin{abstract}

Convolutional layers are the core building blocks of Convolutional Neural Networks (CNNs).
In this paper, we propose to augment a convolutional layer with an additional \emph{depthwise} convolution, where each input channel is convolved with a different 2D kernel.
The composition of the two convolutions constitutes an over-parameterization, since it adds learnable parameters, while the resulting linear operation can be expressed by a single convolution layer.  
We refer to this depthwise over-parameterized convolutional layer as DO-Conv.
We show with extensive experiments that the mere replacement of conventional convolutional layers with DO-Conv layers boosts the performance of CNNs on many classical vision tasks, such as image classification, detection, and segmentation.
Moreover, in the inference phase, the depthwise convolution is folded into the conventional convolution, reducing the computation to be exactly equivalent to that of a convolutional layer without over-parameterization.
As DO-Conv introduces performance gains without incurring any computational complexity increase for inference, we advocate it as an alternative to the conventional convolutional layer.
We open-source a reference implementation of DO-Conv in Tensorflow, PyTorch and GluonCV at \href{https://github.com/yangyanli/DO-Conv}{\underline{https://github.com/yangyanli/DO-Conv}}.
\end{abstract}
\section{Introduction}

Convolutional Neural Networks (CNNs) are capable of expressing highly complicated functions, and have shown great success in addressing many classical computer vision problems, such as image classification, detection and segmentation.
It has been widely accepted that increasing the depth of a network by adding linear and non-linear layers together can increase the network's expressiveness and boost its performance.
On the other hand, adding extra linear layers \emph{only} is not as commonly considered, especially when the additional linear layers result in an \textbf{over-parameterization}\footnote{The meaning of ``over-parameterization'' is overloaded in the community. We opt to use this term and its meaning following~\cite{arora2018optimization}. The term is also used in the community for describing the case where the number of parameters in a model exceeds the size of the training dataset, such as that in~\cite{allen2019convergence}.} --- a case where the composition of consecutive linear layers may be represented by a single linear layer with fewer learnable parameters.

Though over-parameterization does not improve the expressiveness of a network, it has been proven as means of accelerating the training of deep linear networks, and shown empirically to speedup the training of deep non-linear networks~\cite{arora2018optimization}.
These findings suggest that, while much work has been devoted to the quest for novel network architectures, over-parameterization has considerable unexplored potential in benefiting existing architectures.

In this work, we propose to over-parameterize a convolutional layer by augmenting it with an ``extra'' or ``over-parameterizing'' component: a depthwise convolution operation, which convolves separately each of the input channels.
We refer to this depthwise over-parameterized convolutional layer as DO-Conv, and show that it can not only accelerate the training of various CNNs, but also consistently boost the performance of the converged models.

One notable advantage of over-parameterization, in general, is that the multi-layer composite linear operations used by the over-parameterization can be folded into a compact single layer representation after the training phase. Then, only a \emph{single} layer is used at inference time, reducing the computation to be exactly equivalent to a conventional layer.

We show with extensive experiments that using DO-Conv boosts the performance of CNNs on many tasks, such as image classification, detection, and segmentation, merely by replacing the conventional convolution with DO-Conv. Since the performance gains are introduced with \emph{no} increase in inference computations, we advocate DO-Conv as an alternative to the conventional convolutional layer.
\section{Related Work}
The performance of CNNs is highly correlated with their architectures, and a line of notable architectures have been proposed over the recent years, such as AlexNet~\cite{krizhevsky2012imagenet}, VGG~\cite{simonyan2014very}, GoogLeNet \cite{szegedy2015going}, ResNet~\cite{he2016deep}, and MobileNets~\cite{howard2017mobilenets}. Our work is orthogonal to the quest for novel architectures, and can be combined with existing ones to boost their performance.

Convolutional layers are the core building blocks of CNNs. Thus, an improvement of these core building blocks can often lead to boosting the performance of CNNs. Several alternatives to the classical convolutional layer have been proposed, which offer improved feature learning capability and/or efficiency~\cite{dai2017deformable,zhu2019deformable,li2019selective,chen2019drop}. Our work may be viewed as a contribution along this line.

Arora et al.~\cite{arora2018optimization} studied the role of over-parameterization in gradient descent based optimization. They have \emph{proven} that over-parameterization of fully connected layers can accelerate the training of deep linear networks, and shown \emph{empirically} that it can also accelerate the training of deep non-linear networks. There are multiple ways to over-parameterize a layer. The kernel of a convolutional layer has both channel and spatial axes, thus the over-parameterization of a convolutional layer can be more versatile than that of a fully connected layer.
ExpandNets~\cite{guo2018expandnet} proposes to over-parameterize the convolution kernel over the input and output channel axes. The advantage of over-parameterization is demonstrated in \cite{guo2018expandnet} only on the training of compact CNNs, whose performance is no match to that of the mainstream CNNs. Our method over-parameterizes the convolution kernel over its spatial axes, and the advantage is demonstrated on several commonly used CNN architectures.
In ACNet~\cite{ding2019acnet} the convolutional layer is replaced with an Asymmetric Convolution Block. This approach is, in essence, an over-parameterization of the convolution kernel over the center row and column of its spatial axes. In fact, ACNet may be viewed as a special case of our over-parameterization approach, which over-parameterizes the entire kernel.

Over-parameterized layers introduce ``extra'' linear transformations without increasing the expressiveness of the network, and once their parameters have been learned, these ``extra'' transformations are folded in the inference phase. In this sense, normalization layers, such as batch normalization~\cite{ioffe2015batch} and weight normalization~\cite{salimans2016weight}, are quite similar to over-parameterized layers. Normalization layers have been widely used for improving the training of CNNs, while the reason for their success is still being actively studied~\cite{santurkar2018does,bjorck2018understanding,kohler2018towards,kohler2019exponential}. It is intriguing to study whether or not the effectiveness of over-parameterization and normalization in CNNs may be attributed to the same underlying reasons.
\section{Method}
\label{sec:method}

\paragraph{Notation.} We use $\mathbb{T}_{ijk}$ to denote the $(i,j,k)$-th element of a 3D tensor $\mathbb{T} \in R^{I \times J \times K}$. A tensor can be reshaped without changing the number of its elements, or their values, e.g., a 4D tensor $\mathbb{T} \in R^{I \times J \times K \times L}$ can be reshaped to a 3D tensor $\mathbb{T}' \in R^{I \times (J \times K) \times L}$, or $\mathbb{T}' \in R^{I \times M \times L}$, where $M=J \times K$. For simplicity, we will refer to a tensor and its reshaped version(s) with the same symbol. Furthermore, $\mathbb{T}_{ijkl}$ and $\mathbb{T}_{i(jk)l}$ refer to the same element of $\mathbb{T} \in R^{I \times J \times K \times L}$ and its reshaped version $\mathbb{T} \in R^{I \times (J \times K) \times L}$, respectively.

\paragraph{Conventional convolutional layer.} Given an input feature map, a convolutional layer processes it in a sliding window fashion, applying a set of convolution kernels to a patch of corresponding size, at each window position. For the purposes of our exposition, it is convenient to think of a patch as a 2D tensor $\mathbb{P} \in R^{(M \times N) \times C_{in}}$, where $M$ and $N$ are the spatial dimensions of the patch,
\begin{wrapfigure}{h}{0.45\textwidth}
  \begin{center}
    \includegraphics[width=0.45\textwidth]{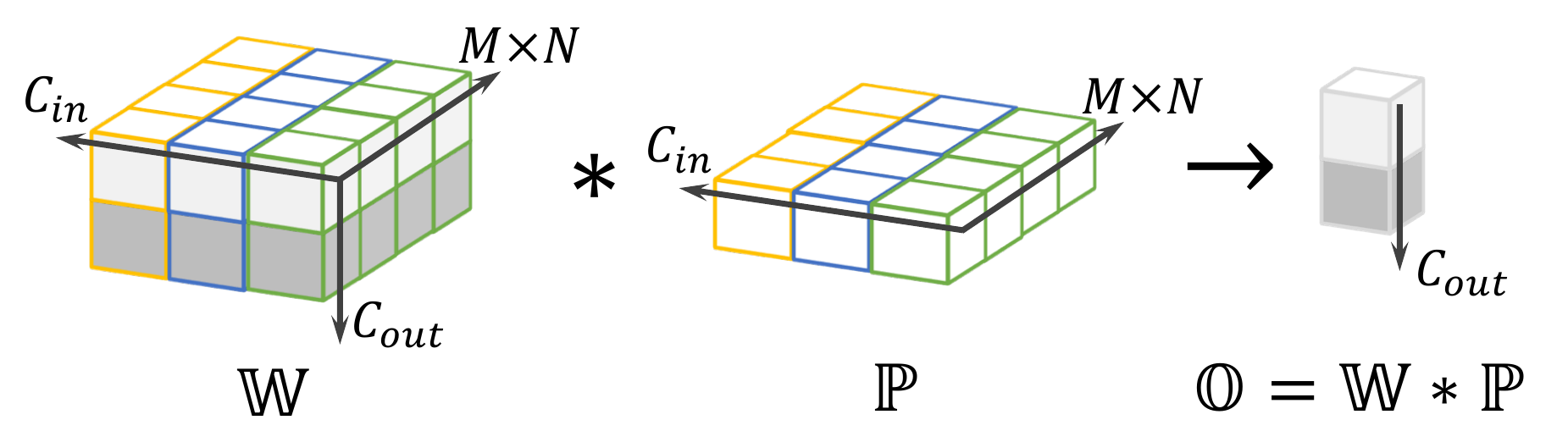}
  \end{center}
  \caption{Conventional convolution.}
  \label{fig:conv}
\end{wrapfigure}
and $C_{in}$ is the number of channels in the input feature map. The trainable kernels of a convolutional layer\footnote{For simplicity, the bias terms are not included in our exposition.} can be represented as a 3D tensor $\mathbb{W} \in R^{C_{out} \times (M \times N) \times C_{in}}$, where $C_{out}$ is the number of channels in the output feature map. The output of a convolution operator $\ast$ is a $C_{out}$-dimensional feature $\mathbb{O} = \mathbb{W} \ast \mathbb{P}$: 
\begin{equation}
     \mathbb{O}_{c_{out}} = \sum_{i}^{(M \times N) \times C_{in}} \mathbb{W}_{c_{out}i} \mathbb{P}_{i}.
    \label{eq:conv}
\end{equation}
This is illustrated in Figure~\ref{fig:conv} (better viewed in color), where $M \times N = 4$, $C_{in} = 3$ with different input channels in different cube frame colors, $C_{out} = 2$ with different output channels in different cube face colors, and each element of $\mathbb{O}$ is produced by a dot-product between one kernel (one horizontal slice of $\mathbb{W}$) and the patch $\mathbb{P}$.

\begin{wrapfigure}{h}{0.15\textwidth}
  \begin{center}
    \includegraphics[width=0.15\textwidth]{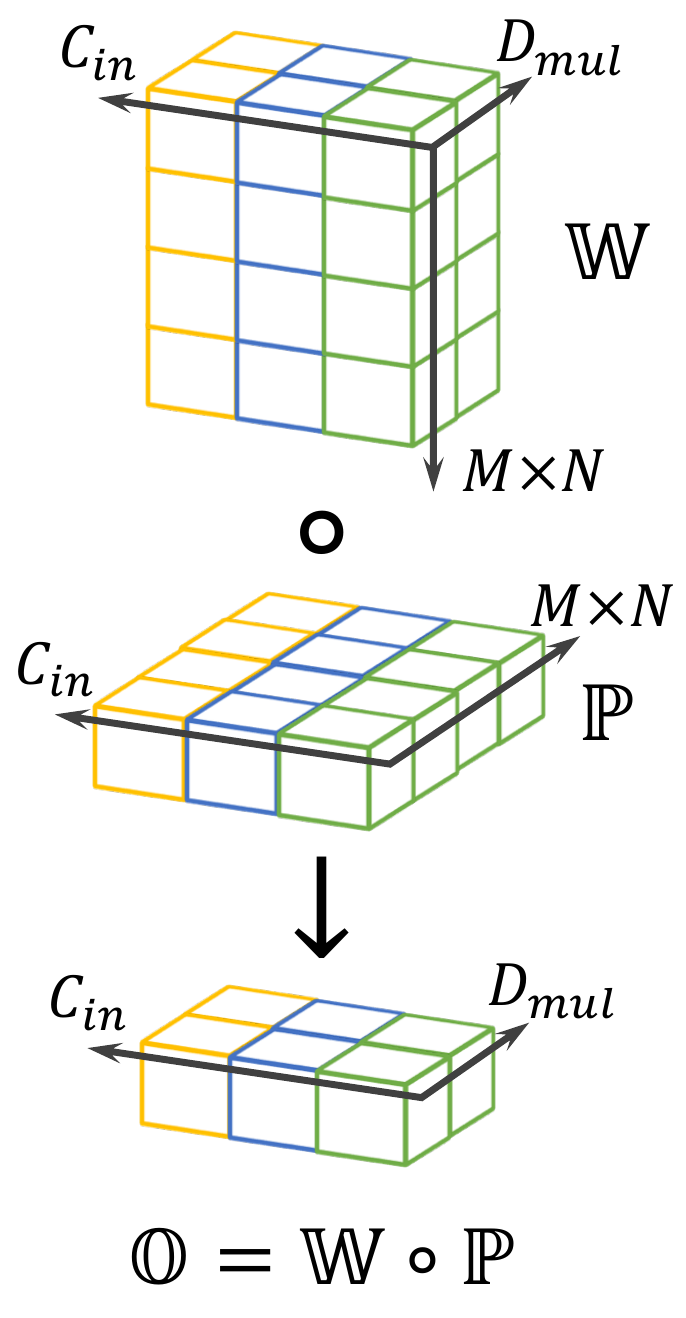}
  \end{center}
  \caption{\\Depthwise\\convolution.}
  \vspace{-1cm}
  \label{fig:dwconv}
\end{wrapfigure}

\paragraph{Depthwise convolutional layer.} In a convolutional layer, dot products are computed between each of the $C_{out}$ kernels and the entire input patch tensor $\mathbb{P}$. In contrast, in depthwise convolution, each of the $C_{in}$ channels of $\mathbb{P}$ is involved in $D_{mul}$ separate dot-products. Thus, each input patch channel (a $M \times N$-dimensional feature) is transformed into a $D_{mul}$-dimensional feature. $D_{mul}$ is often referred to as depth multiplier. As depicted in Figure~\ref{fig:dwconv}, the trainable depthwise convolution kernel can be represented as a 3D tensor $\mathbb{W} \in R^{(M \times N) \times D_{mul} \times C_{in}}$. Since each input channel is converted into a $D_{mul}$-dimensional feature, the output of the depthwise convolution operator $\circ$ is a $D_{mul} \times C_{in}$-dimensional feature $\mathbb{O} = \mathbb{W} \circ \mathbb{P}$:
\begin{equation}
    \mathbb{O}_{d_{mul}c_{in}} = \sum_{i}^{M \times N} \mathbb{W}_{id_{mul}c_{in}} \mathbb{P}_{ic_{in}}.
    \label{eq:dwconv}
\end{equation}
This is illustrated in Figure~\ref{fig:dwconv} (better viewed in color), where $M \times N = 4$, $D_{mul} = 2$, $C_{in} = 3$ with different input channels in different cube frame colors, and each element of $\mathbb{O}$ is computed by the dot product between each vertical column of $\mathbb{W}$ and the elements in the corresponding channel of $\mathbb{P}$ (those with the same color).

\begin{figure*}[t!]	
	\centering
	\includegraphics[width=0.8\textwidth]{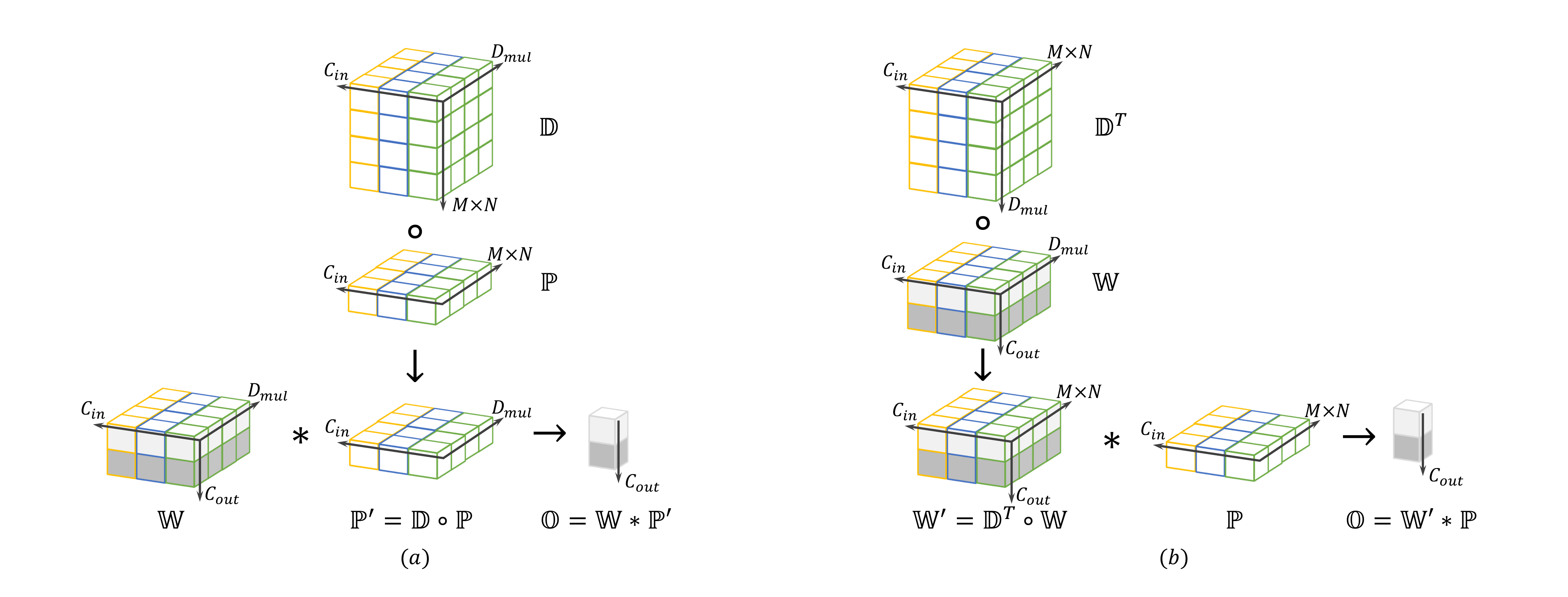}
	\vspace{-0.2cm}
	\caption{DO-Conv operator $\circledast$ over trainable kernels $(\mathbb{D}, \mathbb{W})$, and an input patch $\mathbb{P}$. In this figure, $M \times N = 4$, $D_{mul} = 4$, $C_{in} = 3$ with different input channels illustrated by different cube frame colors, and $C_{out} = 2$ with different output channels illustrated by different cube face colors. DO-Conv operator is a composite of depthwise convolution operator $\circ$ and convolution operator $\ast$, and as depicted in this figure, there are two mathematically equivalent ways for realizing the composition: \emph{feature composition} (a) and \emph{kernel composition} (b). This figure is better viewed in color.
	}
	\label{fig:two-schemes}
\end{figure*}

\paragraph{Depthwise over-parameterized convolutional layer (DO-Conv)} is a composition of a depthwise convolution with trainable kernel $\mathbb{D} \in R^{(M \times N) \times D_{mul} \times C_{in}}$ and a conventional convolution with trainable kernel $\mathbb{W} \in R^{C_{out} \times D_{mul} \times C_{in}}$, \underline{where $D_{mul} \geq M \times N$}. Given an input patch $\mathbb{P} \in R^{(M \times N) \times C_{in}}$, the output of a DO-Conv operator $\circledast$ is, the same as that of a convolutional layer, a $C_{out}$-dimensional feature $\mathbb{O}=(\mathbb{D}, \mathbb{W}) \circledast \mathbb{P}$. More specifically, as illustrated in Figure~\ref{fig:two-schemes}, the depthwise over-parameterized convolution operator can be applied in two mathematically equivalent ways as:
\begin{equation}
    \begin{aligned}
        \mathbb{O}
        &= (\mathbb{D}, \mathbb{W}) \circledast \mathbb{P} \\
        &= \mathbb{W} \ast (\mathbb{D} \circ \mathbb{P}) &&\text{(Fig.~\ref{fig:two-schemes}-a, \emph{feature composition})}\\
        &= (\mathbb{D}^T \circ \mathbb{W}) \ast \mathbb{P}, &&\text{(Fig.~\ref{fig:two-schemes}-b, \emph{kernel composition}),}
    \end{aligned}
    \label{eq:doconv}
\end{equation}
where $\mathbb{D}^T \in R^{D_{mul} \times (M \times N) \times C_{in}}$ is a transpose of $\mathbb{D} \in R^{(M \times N) \times D_{mul} \times C_{in}}$ on the first and second axis. Using \emph{feature composition}, as depicted in Figure~\ref{fig:two-schemes}-a, the depthwise convolution operator $\circ$ first applies the kernel weights $\mathbb{D}$ to the patch $\mathbb{P}$, yielding a transformed feature $\mathbb{P}'=\mathbb{D} \circ \mathbb{P}$, and then a conventional convolution operator $\ast$ applies the kernels $\mathbb{W}$ to  $\mathbb{P}'$, yielding $\mathbb{O}=\mathbb{W} \ast \mathbb{P}'$. In contrast, using \emph{kernel composition}, as depicted in Figure~\ref{fig:two-schemes}-b, the composition of $\circ$ and $\ast$ is achieved through a composite kernel $\mathbb{W}'$, i.e., a depthwise convolution operator first uses the trainable kernels $\mathbb{D}^T$ to transform $\mathbb{W}$, yielding $\mathbb{W}'=\mathbb{D}^T \circ \mathbb{W}$, and then a conventional convolution operator is applied between $\mathbb{W}'$ and $\mathbb{P}$, yielding $\mathbb{O}=\mathbb{W}' \ast \mathbb{P}$.

Note that the receptive field of DO-Conv is still $M \times N$. This is apparent in  Figure~\ref{fig:two-schemes}-b, where the depthwise and conventional convolution kernels are composited to process a single input patch of spatial size $M \times N$. This is perhaps less obvious in the feature composition view (Figure~\ref{fig:two-schemes}-a), where the conventional convolution operator is applied to $\mathbb{P}' \in R^{C_{in} \times D_{mul}}$, while $D_{mul}$ could be greater than $M \times N$. Note, however, that the values of $\mathbb{P}'$ are obtained by transforming an $M \times N$ spatial patch.

\paragraph{DO-Conv is an over-parameterization of convolutional layer.} This statement can be easily verified using the kernel composition formulation (Figure~\ref{fig:two-schemes}-b). The linear transformation performed by a convolutional layer can be concisely parameterized by $C_{out} \times (M \times N) \times C_{in}$ trainable weights. However, in DO-Conv, a linear transformation of equivalent expressiveness is parameterized by two sets of trainable kernel weights: $\mathbb{D} \in R^{(M \times N) \times D_{mul} \times C_{in}}$ and $\mathbb{W} \in R^{C_{out} \times D_{mul} \times C_{in}}$, where $D_{mul} \geq M \times N$. Thus, even for the case where $D_{mul} = M \times N$, the number of parameters is increased by $(M \times N) \times D_{mul} \times C_{in}$.
Note that the condition $D_{mul} \geq M \times N$ is necessary, otherwise the composite operator $\mathbb{W}'$ cannot express the same family of linear transformations as $\mathbb{W}$ in a conventional convolutional layer.

\paragraph{Training and inference of CNNs with DO-Conv.} The interface of DO-Conv is same as that of a convolutional layer, thus DO-Conv layers can easily replace convolutional layers in CNNs. Since the $\circledast$ operator defined in Equation~\ref{eq:doconv} is differentiable, both $\mathbb{D}$ and $\mathbb{W}$ of DO-Conv can be optimized with the gradient descent based optimizer that is used for training CNNs with convolutional layers. After the training phase, $\mathbb{D}$ and $\mathbb{W}$ are folded into $\mathbb{W}'=\mathbb{D}^T \circ \mathbb{W}$, and this single $\mathbb{W}'$ is used for the inference. Since $\mathbb{W}'$ is in the same shape as the kernel of a convolutional layer, the computation of DO-Conv at the inference phase is exactly same as that of a conventional convolutional layer.

\paragraph{Training efficiency and composition choice of DO-Conv.} The two ways for computing DO-Conv are mathematically equivalent, but have different training efficiency. The number of multiply and accumulate operations (MACC), is often used for measuring the amount of computation and serves as an indicator of the efficiency. The MACC cost of feature and kernel composition, when they are applied on an feature map in $R^{H \times W \times C_{in}}$ (assuming $stride=1$), can be calculated as follows:
\begin{equation*}
    \begin{aligned}
        \text{Feature composition} &: \begin{cases} \mathbb{P'} = \mathbb{D} \circ \mathbb{P} &: D_{mul} \times (M \times N) \times C_{in} \times \, $\textcolor{red}{$(H \times W)$}$, \\
         \mathbb{O} = \mathbb{W} \ast \mathbb{P'} &: C_{out} \times C_{in} \times H \times W \times \, $\textcolor{blue}{$D_{mul}$}$, \\
         \end{cases} \\
        \text{Kernel composition} &: \begin{cases}
        \mathbb{W'} = \mathbb{D}^T \circ \mathbb{W} &: D_{mul} \times (M \times N) \times C_{in} \times \, $\textcolor{red}{$C_{out}$}$, \\
        \mathbb{O} = \mathbb{W'} \ast \mathbb{P} &: C_{out} \times C_{in} \times H \times W \times \, $\textcolor{blue}{$(M \times N)$}$, \\
        \end{cases}
    \end{aligned}
\end{equation*}
where $H$ and $W$ are the height and width of the feature map, respectively. Note that $\mathbb{W'}$ is computed once for an entire feature map. The MACC costs of feature and kernel composition depend on the values of the involved hyper-parameters. However, in most cases, since $H \times W \gg C_{out}$ and $D_{mul} \geq M \times N$, kernel composition typically incurs fewer MACC operations than feature composition. Similarly, the memory consumed by $\mathbb{W'}$ in kernel composition is typically smaller than that consumed by $\mathbb{P'}$ in feature composition. Therefore, kernel composition is preferable for the training phase.

\paragraph{DO-Conv and depthwise separable convolutional layer~\cite{chollet2017xception}.} The feature composition of DO-Conv (Figure~\ref{fig:two-schemes}-a) is equivalent to applying a $M \times N$ depthwise convolutional layer on an input feature map, yielding an intermediate feature map in $D_{mul} \times C_{in}$ channels, and then applying a $1 \times 1$ convolutional layer on the intermediate feature map. This process is exactly same as that of a depthwise separable convolutional layer, where the $1 \times 1$ convolution is often referred to as pointwise convolution. However, the motivation of DO-Conv and depthwise separable convolution is quite different. Depthwise separable convolution is introduced as an approximation and alternative to a conventional convolution for saving MACC to ease the deployment of CNNs especially on edge devices, thus it requires\footnote{In earlier versions of Tensorflow~\cite{tensorflow2015-whitepaper}, setting $D_{mul} > M \times N$ in the depthwise separable convolutional layer is considered as an error, though this constraint has been removed since commit \href{https://github.com/tensorflow/tensorflow/commit/1822073137e1ac431250ea6f89b2719aac8d4782}{\underline{1822073}}.} $D_{mul} < M \times N$, and $D_{mul}=1$ is probably the choice most widely used in practice, for example in Xception~\cite{chollet2017xception} and MobileNets~\cite{howard2017mobilenets,sandler2018mobilenetv2,howard2019searching}.

\begin{wrapfigure}{r}{0.5\textwidth}
  \begin{center}
  \vspace{-0.5cm}
    \includegraphics[width=0.45\textwidth]{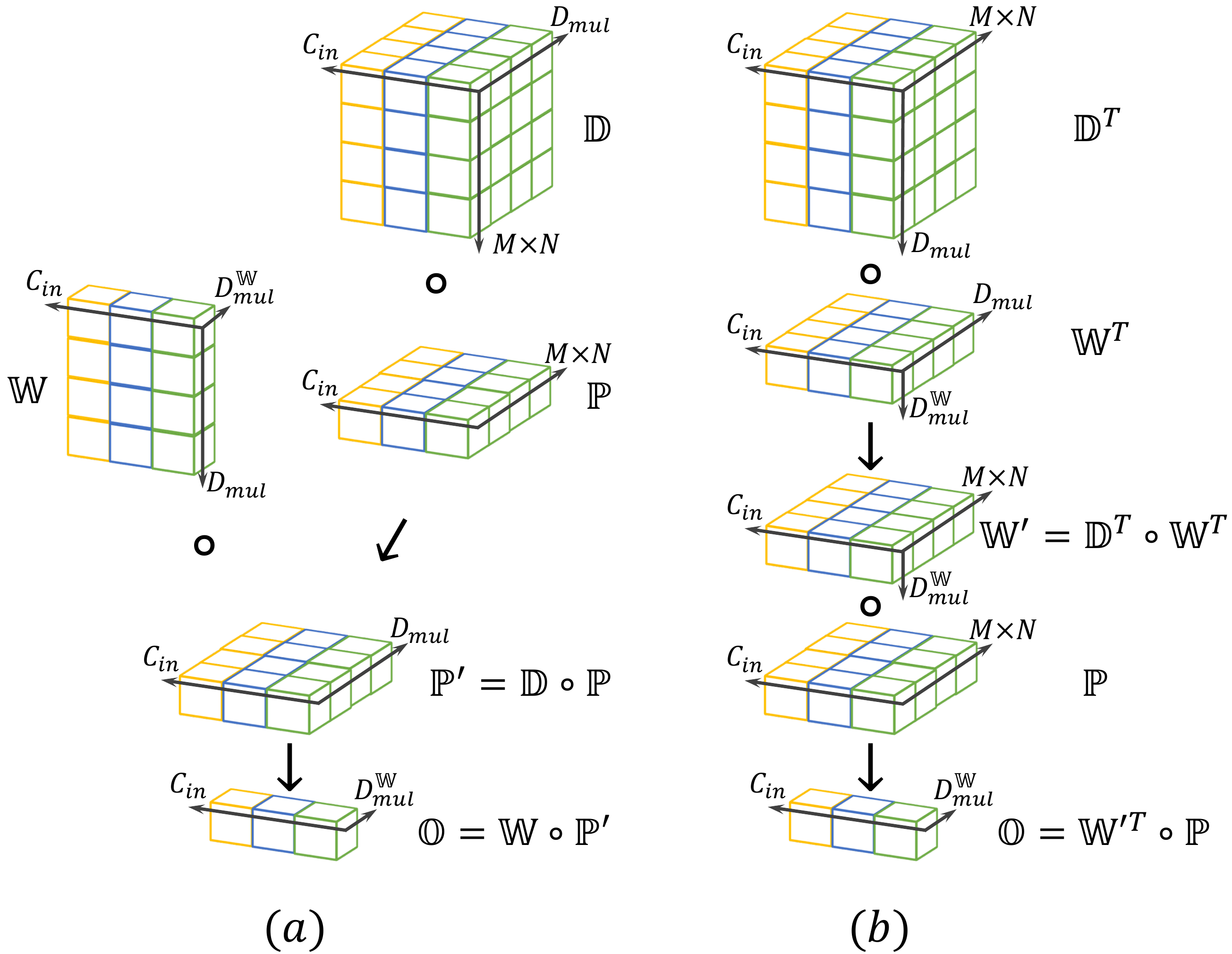}
  \end{center}
  \vspace{-0.3cm}
  \caption{DO-DConv operator $\circledcirc$ over trainable kernels $(\mathbb{D}, \mathbb{W})$, and an input patch $\mathbb{P}$. In this figure, $M \times N = 4$, $D_{mul} = 4$, $D^{\mathbb{W}}_{mul} = 1$, $C_{in} = 3$ with different input channels indicated by different cube frame colors. DO-DConv operator is a composite of two depthwise convolution operators, and similar to that in Figure~\ref{fig:two-schemes}, there are two mathematically equivalent ways for realizing the composition: \emph{feature composition} (a) and \emph{kernel composition} (b). This figure is better viewed in color.}
  \vspace{-5mm}
  \label{fig:DO-DConv}
\end{wrapfigure}

\paragraph{Training depthwise separable convolutional layer with kernel composition.} The equivalence between feature and kernel composition holds not only for $D_{mul} \geq M \times N$ (DO-Conv), but also for $D_{mul} < M \times N$ (depthwise separable convolution). We have shown that for DO-Conv, kernel composition often saves MACC and memory, compared with feature composition. For depthwise separable convolution, it is easy to see that feature composition is more economical. While the MACC is highly correlated with the running speed on edge devices that are often computation capability bounded, this might not be the case on NVIDIA GPUs, as they are more memory access bounded. Depthwise separable convolution has a significantly lower MACC cost than a conventional convolution, but it often runs slower on NVIDIA GPUs, which the CNNs are often trained on. To speedup the training of depthwise separable convolution layers on NVIDIA GPUs, kernel composition can be used, after which feature composition can be used for the deployment on edge devices. \emph{This handy trick is not the focus of this paper, but a by-product of the feature and kernel composition equivalence that may greatly interest many deep learning practitioners.}

\paragraph{Depthwise over-parameterized depthwise/group convolutional layer (DO-DConv/DO-GConv).} Depthwise over-parameterization can not only be applied over a conventional convolution to yield DO-Conv, but also be applied over a depthwise convolution, which leads to DO-DConv. Following the same principle we used for establishing DO-Conv, as shown in Figure~\ref{fig:DO-DConv}, DO-DConv also can be computed in two mathematically equivalent ways as:
\begin{equation}
    \begin{aligned}
        \mathbb{O}
        &= (\mathbb{D}, \mathbb{W}) \circledcirc \mathbb{P} \\
        &= \mathbb{W} \circ (\mathbb{D} \circ \mathbb{P}) &&\text{(Fig.~\ref{fig:DO-DConv}-a, \emph{feature composition})}\\
        &= (\mathbb{D}^T \circ \mathbb{W}^T)^T \circ \mathbb{P}, &&\text{(Fig.~\ref{fig:DO-DConv}-b, \emph{kernel composition})}
    \end{aligned}
    \label{eq:do-dconv}
\end{equation}
DO-DConv layer can be used as an alternative to depthwise convolutional layer, which is widely used in MobileNets~\cite{howard2017mobilenets,sandler2018mobilenetv2,howard2019searching}. Same as in DO-Conv, both $\mathbb{D}$ and $\mathbb{W}$ are used in the training of DO-DConv, while the folded $\mathbb{W'}$ is used for the inference. Following the same principle used for yielding DO-Conv and DO-DConv, grouped convolutional layer~\cite{krizhevsky2012imagenet}, which is a general case of depthwise convolutional layer and is shown to be effective in ResNeXt~\cite{xie2017aggregated}, can be over-parameterized, yielding DO-GConv. We refer to the over-parameterized (depthwise/group) convolutional layer as DO-Conv for simplicity, in cases where there is no ambiguity.
\section{Experiments}
\label{sec:experiments}

\paragraph{The choice of $D_{mul}$ and initialization of $\mathbb{D}$.} Note that when $M \times N = 1$, the underlying convolution is a pointwise convolution. In this case, there is no spatial patch for the application of the depthwise convolution, thus DO-Conv is not applied. When $D_{mul}=M \times N$, the kernel of the underlying convolution $\mathbb{W}$ is of the same shape as the conventional one, and the over-parameterizing kernel $\mathbb{D} \in R^{(M \times N) \times D_{mul} \times C_{in}}$ can be viewed as $C_{in}$ square $D_{mul} \times D_{mul}$ matrices. We initialize these square matrices to identity $\mathbb{I}$, such that $\mathbb{W}'=\mathbb{W}$ in the beginning. In this case, the over-parameterization does not initially change the operation of the original CNN. It also facilitates DO-Conv by reusing the parameters from the pre-trained model. Moreover, since the diagonal elements in $\mathbb{D}$ could be overly suppressed by $L_2$ regularization, we optimize $\mathbb{D}'=\mathbb{D}-\mathbb{I}$, where $\mathbb{D}'$ is initialized with zeros, instead of optimizing $\mathbb{D}$ directly. Unless otherwise noted, we use $D_{mul}=M \times N$ in our experiments.

\paragraph{Comparison protocol.} The performance of a CNN can be affected by a wide range of factors. To demonstrate the effectiveness of DO-Conv, we take several notable CNNs as baselines, and merely replace the non-pointwise convolutional layers with DO-Conv, \emph{without any change to other settings}. In other words, the replacement is the one and only difference between the baseline and our method. This guarantees that the observed performance change is due to the application of DO-Conv, but not other factors. Furthermore, this also means that no hyper-parameter is tuned to favor DO-Conv. Following this protocol, we demonstrate the effectiveness of DO-Conv on image classification, semantic segmentation and object detection tasks on several benchmark datasets.

\subsection{Image Classification}

We conducted image classification experiments on CIFAR~\cite{krizhevsky2009learning} and ImageNet~\cite{russakovsky2015imagenet}, with a set of notable architectures including ResNet-v1~\cite{he2016deep} (including ResNet-v1b, which modifies ResNet-v1 by setting the $2 \times 2$ stride at the $3 \times 3$ layer of a bottleneck block), Plain (same as ResNet-v1, but without skip links), ResNet-v2~\cite{he2016identity}, ResNeXt~\cite{xie2017aggregated},  MobileNet-v1~\cite{howard2017mobilenets}, -v2~\cite{sandler2018mobilenetv2} and -v3~\cite{howard2019searching}.

\begin{wraptable}{r}{8.5cm}
  \centering
  \scalebox{0.56}{
      \begin{tabular}{ c |c| c | c|c  |c| c | c|c|c| c}
      \hline
         \multicolumn{2}{c|}{Network} & Plain &\multicolumn{4}{c|}{ResNet-v1} & \multicolumn{4}{c}{ResNet-v2}\\
         \cline{1-11} 
         \multicolumn{2}{c|}{Depth} &20 &20 &56&110&164&20 & 56&110&164\\
         \cline{1-11} 
         CIFAR &Baseline &90.88&92.03  &93.47 &94.00 &94.65 &91.83 &93.26  &93.72&94.69\\
         \cline{2-11}
         -10 &DO-Conv  &\textcolor{red}{+0.37} &\textcolor{red}{+0.25}   &\textcolor{red}{+0.05} &\textcolor{red}{-0.05} &\textcolor{red}{-0.06}  &\textcolor{red}{+0.39}  &\textcolor{red}{+0.12}  &\textcolor{red}{+0.21} &\textcolor{red}{-0.07} \\
         \cline{1-11}
        CIFAR &Baseline &66.01&67.37 &70.65 &72.58  &75.43   &67.14  &70.15 &72.00 &75.32 \\
         \cline{2-11}
        -100&DO-Conv  &\textcolor{red}{+0.27}  &\textcolor{red}{+0.31} &\textcolor{red}{+0.25} &\textcolor{red}{+0.01}&\textcolor{red}{+0.03} &\textcolor{red}{+0.54} &\textcolor{red}{+0.63} &\textcolor{red}{+0.22}   &\textcolor{red}{+0.07} \\
         \hline
      \end{tabular}
  }
  \caption{
  \label{tab:cifar_do_conv}
  Top-1 classification accuracy (\%) on CIFAR.}
\end{wraptable}

The experiments on CIFAR follow the same settings as those in~\cite{he2016deep,he2016identity} and the results are shown in Table~\ref{tab:cifar_do_conv}, where the ``DO-Conv'' rows show the performance change relative to the baselines. All the results on CIFAR dataset reported in this paper are the averaged accuracy of the last five epochs over five runs. We can observe that DO-Conv brings a promising improvement over the baselines on relatively shallower networks.

\begin{table*}[ht]
  \centering
  \scalebox{0.62}{
      \begin{tabular}{c | c | c| c|c|c|c | c| c| c|c|c |c |c | c| c|c}
      \hline 
        Network &Plain &\multicolumn{5}{c|}{ResNet-v1}   &\multicolumn{5}{c|}{ResNet-v1b}   &\multicolumn{5}{c}{ResNet-v2}\\
         \cline{1-17} 
         Depth &18 &18 &34 &50 &101 &152 &18 &34 &50 &101 &152 &18 &34 &50 &101 &152 \\
          \cline{1-17} 
         Reference &-  &70.93 & 74.37  & 77.36& 78.34 &79.22& 70.94 &74.65&77.67  &79.20 &79.69&71.00 &74.40  &77.17 &78.53 &79.21\\
           \cline{1-17} 
          Baseline &69.97 &70.87 &74.49 &77.32 &78.16 &79.34&71.08&74.35 &77.56 &79.14  &79.60&70.80  &74.76  &77.17  &78.56&79.24  \\
           \cline{1-17} 
          DO-Conv   & \textcolor{red}{+1.01} &\textcolor{red}{+0.82}&\textcolor{red}{+0.49} &\textcolor{red}{+0.08} &\textcolor{red}{+0.46}&\textcolor{red}{+0.07}&\textcolor{red}{+0.71} &\textcolor{red}{+0.77} &\textcolor{red}{+0.44}  &\textcolor{red}{+0.25}  &\textcolor{red}{+0.1}  &\textcolor{red}{+0.64}&\textcolor{red}{+0.22}&\textcolor{red}{+0.31}&\textcolor{red}{+0.11}&\textcolor{red}{+0.14}\\
         \hline
      \end{tabular}
  }
  \caption{
  \label{tab:imagenet_do_conv}
  Top-1 classification accuracy (\%) on ImageNet.}
\end{table*}

\begin{wraptable}{r}{7cm}
\centering
\vspace{-0.05\linewidth}
  \scalebox{0.8}{
      \begin{tabular}{ c |c| c | c|c }
      \hline 
        \multirow{2}{*}{Network} &\multicolumn{3}{c|}{MobileNet}& ResNeXt\\
        \cline{2-4}
        &v1 & v2 & v3 &50$\_$32x4d\\
        \cline{1-5} 
        Reference &73.28  &72.04 &75.32  &79.32\\
        \cline{1-5} 
        Baseline  &73.30  &71.89  &75.16 &79.21\\
        \cline{1-5} 
        DO-D/GConv       &\textcolor{red}{+0.03}&\textcolor{red}{+0.16}&\textcolor{red}{+0.14} &\textcolor{red}{+0.40}   \\
        \hline
        \end{tabular}
    }
    \caption{
    \label{tab:imagenet_do_gconv}
    Top-1 classification accuracy (\%) on ImageNet with DO-D/GConv (MobileNets/ResNeXt).}
\end{wraptable}

 The experiments on ImageNet follow the same settings as those in the model zoo of GluonCV~\cite{guo2020gluoncv}. We implemented DO-Conv into GluonCV (commit \href{https://github.com/dmlc/gluon-cv/tree/bbe41662e877ede84ab84c403654a8a87366b747}{\underline{bbe4166}}), such that it is convenient to make sure DO-Conv is the only change over baselines, since all of the other settings are not touched. We base the experiments on GluonCV since it provides not only a wide variety of high performance CNNs, but also the their training procedures. We consider GluonCV highly reproducible, but still, to exclude clutter factors as much as possible, we train these CNNs as baseline ourselves, and compare DO-Conv versions with them, while reporting the performance provided by GluonCV as reference. The results of DO-Conv and DO-DConv/DO-GConv are summarized in Table~\ref{tab:imagenet_do_conv} and Table~\ref{tab:imagenet_do_gconv}, respectively. It is clear that DO-Conv consistently boosts the performance of various baselines on ImageNet classification.

\begin{wraptable}{r}{7cm}
    \centering
    \vspace{-1.2cm}
    \scalebox{0.8}{
    \begin{tabular}{c|c|c|c}
    \hline
         \multicolumn{2}{c|}{Training Stage} &\multicolumn{2}{c}{Mean IoU.(\%)}  \\
         \hline
         Backbone &Segmentation &PASCAL VOC &Cityscapes\\
         \hline
         Baseline & Baseline & 88.59  &78.71\\
         \hline
         Baseline & DO-Conv &\textcolor{red}{+0.25}  &\textcolor{red}{+1.45}\\
         \hline
         DO-Conv  & DO-Conv  &\textcolor{red}{+0.05} &\textcolor{red}{+0.82}\\
         \hline
         
    \end{tabular}
    }
    \caption{\label{tab:segmentation}
    Semantic segmentation performance on PASCAL VOC and Cityscapes.}
\end{wraptable}

\subsection{Semantic Segmentation}
We also conducted semantic segmentation experiments on the PASCAL VOC~\cite{everingham2015pascal} and the Cityscapes datasets~\cite{cordts2016cityscapes} with GluonCV, following the same comparison protocol as that used for the ImageNet classification task. Differently from image classification, the training of a CNN model for image segmentation often has two stages: the ``Backbone'' and ``Segmentation''. The first stage pre-trains a backbone model on the ImageNet classification task, and the second stage fine-tunes the backbone for the segmentation task. DO-Conv can be used in one or both of the stages. We use Deeplabv3~\cite{chen2017rethinking} with ResNet-50 and ResNet-100 as the backbones for Cityscapes and PASCAL VOC datasets, respectively, and summarize the results in Table~\ref{tab:segmentation}.
Note that the delta in both the second and third rows are relative to the first row. We can observe that DO-Conv consistently boosts the performance, either when used only in the second stage, or in both stages.

\begin{wraptable}{r}{7cm}
  \centering
  \vspace{-1.2cm}
  \scalebox{0.65}{
    \begin{tabular}{c|c |  c|c |  c|c |  c|c}
       \hline 
         \multicolumn{2}{c|}{Training Stage}   &\multirow{2}{*}{$AP$} 
         &\multirow{2}{*}{$AP_{50}$} &\multirow{2}{*}{$AP_{75}$} &\multirow{2}{*}{$AP_{S}$} &\multirow{2}{*}{$AP_{M}$} &\multirow{2}{*}{$AP_{L}$} \\
         \cline{1-2}
         Backbone & Detection & & & & & & \\
        \hline
        Baseline & Baseline  &37.9 &59.5 &41.2 &21.4 &41.4 &49.6\\
        \hline
        Baseline & DO-Conv
         & \textcolor{red}{+0.0}&	\textcolor{red}{+0.4}&	\textcolor{red}{-0.2}	&\textcolor{red}{+0.7}&	\textcolor{red}{+0.6}	&\textcolor{red}{-1.0}\\
        \hline
        
         DO-Conv & DO-Conv      
        &\textcolor{red}{+0.3}	&\textcolor{red}{+0.4}	&\textcolor{red}{+0.3}	&\textcolor{red}{+1.0}  &\textcolor{red}{+0.3}	&\textcolor{red}{+0.0}\\
        \hline
    \end{tabular}
    }
  \caption{
    \label{tab:detection}
    Object detection performance (\%) on COCO. We followed the same standard metrics as in ~\cite{lin2017focal}, i.e., $AP$ denotes the Average Precision at  IoU=0.50:0.05:0.95 which is a primary challenge metric, $AP_{50}$ and $AP_{75}$ are AP at IoU=0.50, 0.75, respectively, $AP_{S}$, $AP_{M}$ and $AP_{L}$ denote the $AP$ for small, medium and large objects, respectively.}
\end{wraptable}

\subsection{Object Detection}
We evaluated DO-Conv for object detection task on the COCO~\cite{lin2014microsoft} dataset using Faster R-CNN~\cite{Ren2015FasterRT} with ResNet-50~\cite{he2016deep} backbone, again with GluonCV, following the same aforementioned comparison protocol. The results are summarized in Table~\ref{tab:detection}. Similarly to segmentation, the detection task has two stages: the ``Backbone'' and ``Detection''. Note that the delta in both the second and third rows are relative to the first row. We can observe that using DO-Conv in the ``Detection'' stage only does not improve the overall performance. Yet, when it is used in both stages, an obvious improvement is achieved. Remember that no hyper-parameter tuning was done when making these comparisons, and all the experiments stick to the hyper-parameters tuned for the baselines.

\subsection{Visualizations}

\begin{figure*}[h!]	
	\centering
	\includegraphics[width=\textwidth]{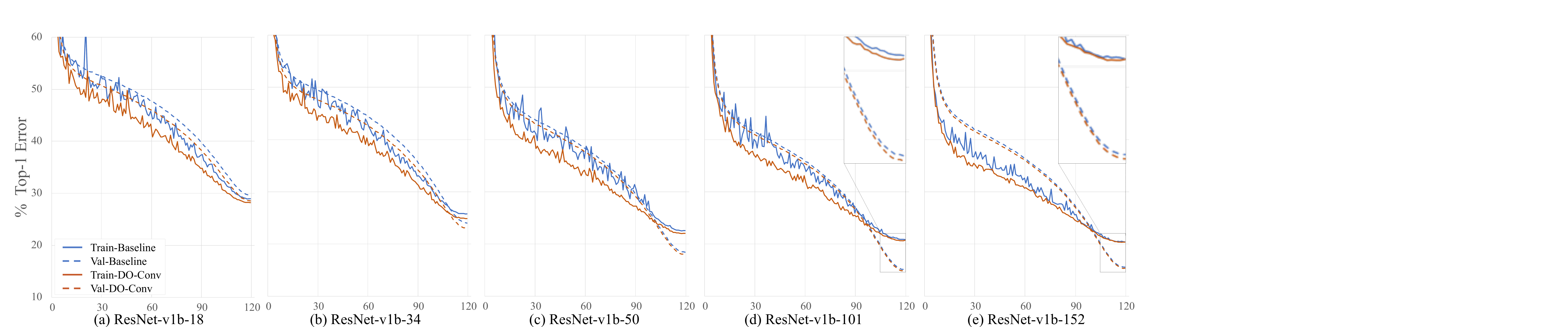}
	\vspace{-0.5cm}
	\caption{Train and validation curves on ImageNet classification with ResNet-v1b in different depths.}
	\label{fig:train_val_curves}
\end{figure*}

\vspace{-0.5cm}
\paragraph{Training dynamics.} We show the train and validation curves of baseline and DO-Conv over 120 epochs with ResNet-v1b in different depths in Figure~\ref{fig:train_val_curves}. The hyper-parameters of ResNet in GluonCV are different from those used in the original ResNet work~\cite{he2016deep}. One notable difference is that the learning rate is gradually decayed to zero at the end of training, and the training converges to a significantly better performance than that reported in \cite{he2016deep}. It is clear that the training of DO-Conv not only converges faster, but also converges to lower errors. While faster convergence due to over-parameterization has been reported before \cite{arora2018optimization}, to the best of our knowledge, we are the first to report convergence to lower errors on mainstream architectures.

\begin{figure*}[h!]
	\centering
	\includegraphics[width=0.9\textwidth]{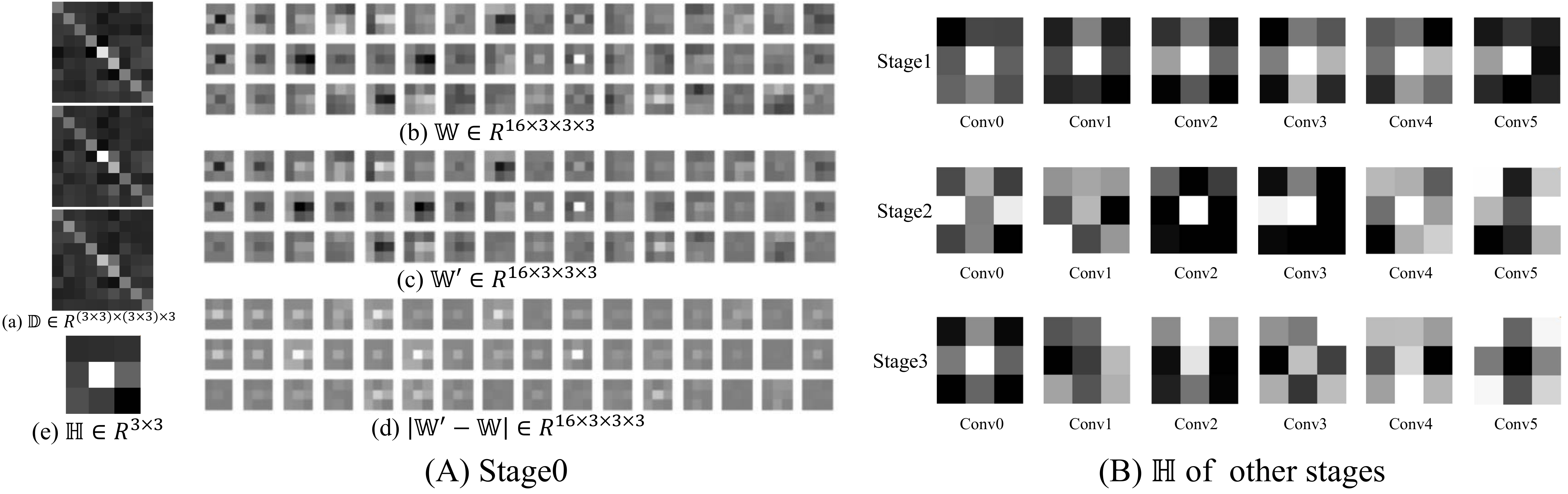}
	\caption{Visualization of ResNet-v2-20 kernels trained from CIFAR-100 classification. For the convenience of observation, each visualization is normalized, with the brighter/darker colors correspond to the larger/smaller values, respectively.}
	\label{fig:kernel_s}
\end{figure*}

\paragraph{Effect of $\mathbb{D}$ on $\mathbb{W}$.} This is visualized in Figure~\ref{fig:kernel_s}, where $\mathbb{H}$ is the accumulated absolute difference between $\mathbb{W'}$ and $\mathbb{W}$, i.e., $\mathbb{H}_{mn}=\sum_{i}^{C_{out} \times C_{in}} |\mathbb{W'}_{imn}-\mathbb{W}_{imn}|$. We found that while some $\mathbb{H}$s exhibit strong skeleton pattern (major differences on center row and column) as observed in ACNet~\cite{ding2019acnet}, other $\mathbb{H}$s exhibit rather different patterns, such as ``Stage2 Conv5'' and ``Stage3 Conv5'' in Figure~\ref{fig:kernel_s} (B).

\subsection{Ablation Studies}

\paragraph{DO-Conv in different ResNet stages.}
\begin{wraptable}{r}{6cm}
	\centering
	\vspace{-2.0cm}
	\scalebox{0.65}{
	\begin{tabular}{ c c c c c| c |c}
	\hline 
         \multicolumn{5}{c|}{Stage of ResNet} &CIFAR-100&ImageNet \\
       \cline{1-5}
        0 &1 &2 &3 &4 &ResNet-v2-20 &ResNet-v1b-50\\
        \cline{1-7}
         \multicolumn{5}{c|}{Baseline}  &67.14  &77.56  \\
       \cline{1-7}  
        $\checkmark$ & & & & &\textcolor{red}{+0.17} &\textcolor{red}{+0.17} 
        \\
        \cline{1-7} 
         &$\checkmark$ & & & &\textcolor{red}{+0.02} &\textcolor{red}{-0.07} \\
        \cline{1-7}
         & &$\checkmark$ & & &\textcolor{red}{+0.25} &\textcolor{red}{-0.03}\\
        \cline{1-7}
         & & &$\checkmark$ & &\textcolor{red}{+0.29} &\textcolor{red}{+0.02}\\
        \cline{1-7}
        & & & & $\checkmark$&- &\textcolor{red}{+0.06}\\
        \cline{1-7}
        $\checkmark$ &$\checkmark$ & & & & \textcolor{red}{+0.10} &\textcolor{red}{-0.11}\\
        \cline{1-7}
        $\checkmark$ &$\checkmark$ &$\checkmark$ & & &\textcolor{red}{+0.32} &\textcolor{red}{+0.16}\\
        \cline{1-7}
        $\checkmark$ &$\checkmark$ &$\checkmark$ &$\checkmark$ & &\textcolor{red}{+0.55} &\textcolor{red}{+0.14}\\
        \cline{1-7}
        $\checkmark$ &$\checkmark$ &$\checkmark$ &$\checkmark$ &$\checkmark$ &- &\textcolor{red}{+0.44}\\
        \hline
	\end{tabular}
	}
	\caption{\label{tab:specific ab}
	Top-1 accuracy (\%) with DO-Conv in different stages. The deltas are relative to the baseline.}
\end{wraptable}
We show the results of replacing the convolutional layers in subsets of ResNet stages with DO-Conv in Table~\ref{tab:specific ab}. For ResNet-v2-20 on CIFAR-100, the accuracy consistently improves when more DO-Conv are used. For ResNet-v1b-50 on ImageNet, the effect is more complicated, as the use of DO-Conv in certain stages may result in a performance drop. The effectiveness of DO-Conv in the first stage is consistent in both cases.

\paragraph{Different initialization of $\mathbb{D}$.}
\begin{wraptable}{r}{6cm}
	\centering
	\vspace{-1.cm}
	\scalebox{0.8}{
	\begin{tabular}{ c |c }
	\hline 
        Baseline  &77.56   \\
       \hline 
        DO-Conv (random-init)  &\textcolor{red}{+0.18}\\
        \hline
        DO-Conv (identity-init)       &\textcolor{red}{+0.44} \\
        \hline
	\end{tabular}
	}
	\caption{\label{tab:diff_init}
	Top-1 accuracy (\%) on ImageNet with ResNet-v1b-50 in different initializations. The deltas relative to the baseline.}
\end{wraptable}
The kernels in CNNs are usually randomly initialized when training the networks from scratch. Since the role of the over-parameterization kernel is different than that of other kernels, we opt to initialize it as identity. However, we found that even if it is randomly initialized (with He initialization~\cite{he2015delving}), it can still boost the performance, though the gain is smaller, compared to identity initialization, as shown in Table~\ref{tab:diff_init}.

\paragraph{What if $D_{mul} > M \times N$?}
\begin{wraptable}{r}{6cm}
	\centering
	\vspace{-1.cm}
	\scalebox{0.7}{
	\begin{tabular}{ c |c }
	\hline 
        Baseline  &67.14   \\
       \hline 
        DO-Conv ($D_{mul}=M \times N$)  &\textcolor{red}{+0.55}\\
        \hline
        DO-Conv ($D_{mul}=(M+2) \times (N+2)$)      &\textcolor{red}{+0.36} \\
        \hline
        DO-Conv ($D_{mul}=(M\times2) \times (N\times2)$)     &\textcolor{red}{+0.47} \\
        \hline
	\end{tabular}}
	\caption{\label{tab:diff_dmul}
	Top-1 accuracy (\%) on CIFAR-100 with ResNet-v2-20 in different $D_{mul}$s. The deltas are relative to the baseline.}
\end{wraptable}
In this case, the $C_{in}$ matrices of $\mathbb{D}$ are no longer square, and they are initialized by filling $(M \times N) \times (M \times N)$ identity matrices as much as possible, while the remaining elements are initialized to zeros. We tested two such settings of $D_{mul}$ and summarized the results in Table~\ref{tab:diff_dmul}. We can observe that all three choices of $D_{mul}$ yield an improvement over the baseline. However, the performance using $D_{mul}=M \times N$ is the best among the three, thus we empirically use $D_{mul}=M \times N$.

\section{Conclusions and Future work}
\label{sec:conclusion}

DO-Conv, a depthwise over-parameterized convolutional layer, is a novel, simple and generic way for boosting the performance of CNNs. Beyond the practical implications of improving training and final accuracy for existing CNNs, without introducing extra computation at the inference phase, we envision that the unveiling of its advantages could also encourage further exploration of over-parameterization as a novel dimension in network architecture design.

In the future, it would be intriguing to get a theoretical understanding of this rather simple means in achieving the surprisingly non-trivial performance improvements on a board range of applications. Furthermore, we would like to expand the scope of applications where these over-parameterized convolution layers may be effective, and learn what hyper-parameters can benefit more from it.

\section*{Broader Impact}

We believe that the impact of our work is mainly to improve the performance of existing CNN models on a variety of computer vision tasks. It could also assist in developing CNN-based solutions to tasks which have not yet been tackled in this manner. We do not believe that our work has any ethical or social implications beyond those stated above.

%
%
\bibliographystyle{splncs04}
\bibliography{main}

\begin{thebibliography}{10}
\providecommand{\url}[1]{\texttt{#1}}
\providecommand{\urlprefix}{URL }
\providecommand{\doi}[1]{https://doi.org/#1}

\bibitem{tensorflow2015-whitepaper}
Abadi, M., Agarwal, A., Barham, P., Brevdo, E., Chen, Z., Citro, C., Corrado,
  G.S., Davis, A., Dean, J., Devin, M., Ghemawat, S., Goodfellow, I., Harp, A.,
  Irving, G., Isard, M., Jia, Y., Jozefowicz, R., Kaiser, L., Kudlur, M.,
  Levenberg, J., Man\'{e}, D., Monga, R., Moore, S., Murray, D., Olah, C.,
  Schuster, M., Shlens, J., Steiner, B., Sutskever, I., Talwar, K., Tucker, P.,
  Vanhoucke, V., Vasudevan, V., Vi\'{e}gas, F., Vinyals, O., Warden, P.,
  Wattenberg, M., Wicke, M., Yu, Y., Zheng, X.: {TensorFlow}: Large-scale
  machine learning on heterogeneous systems (2015),
  \url{http://tensorflow.org/}, software available from tensorflow.org

\bibitem{allen2019convergence}
Allen-Zhu, Z., Li, Y., Song, Z.: A convergence theory for deep learning via
  over-parameterization. In: International Conference on Machine Learning. pp.
  242--252 (2019)

\bibitem{arora2018optimization}
Arora, S., Cohen, N., Hazan, E.: On the optimization of deep networks: Implicit
  acceleration by overparameterization. In: 35th International Conference on
  Machine Learning (2018)

\bibitem{bjorck2018understanding}
Bjorck, N., Gomes, C.P., Selman, B., Weinberger, K.Q.: Understanding batch
  normalization. In: Advances in Neural Information Processing Systems. pp.
  7694--7705 (2018)

\bibitem{chen2017rethinking}
Chen, L.C., Papandreou, G., Schroff, F., Adam, H.: Rethinking atrous
  convolution for semantic image segmentation. arXiv preprint arXiv:1706.05587
  (2017)

\bibitem{chen2019drop}
Chen, Y., Fan, H., Xu, B., Yan, Z., Kalantidis, Y., Rohrbach, M., Yan, S.,
  Feng, J.: Drop an octave: Reducing spatial redundancy in convolutional neural
  networks with octave convolution. In: Proceedings of the IEEE International
  Conference on Computer Vision. pp. 3435--3444 (2019)

\bibitem{chollet2017xception}
Chollet, F.: Xception: Deep learning with depthwise separable convolutions. In:
  Proceedings of the IEEE Conference on Computer Vision and Pattern
  Recognition. pp. 1251--1258 (2017)

\bibitem{cordts2016cityscapes}
Cordts, M., Omran, M., Ramos, S., Rehfeld, T., Enzweiler, M., Benenson, R.,
  Franke, U., Roth, S., Schiele, B.: The cityscapes dataset for semantic urban
  scene understanding. In: Proceedings of the IEEE Conference on Computer
  Vision and Pattern Recognition. pp. 3213--3223 (2016)

\bibitem{dai2017deformable}
Dai, J., Qi, H., Xiong, Y., Li, Y., Zhang, G., Hu, H., Wei, Y.: Deformable
  convolutional networks. In: Proceedings of the IEEE International Conference
  on Computer Vision. pp. 764--773 (2017)

\bibitem{ding2019acnet}
Ding, X., Guo, Y., Ding, G., Han, J.: {ACNet}: Strengthening the kernel
  skeletons for powerful {CNN} via asymmetric convolution blocks. In:
  Proceedings of the IEEE International Conference on Computer Vision. pp.
  1911--1920 (2019)

\bibitem{everingham2015pascal}
Everingham, M., Eslami, S.A., Van~Gool, L., Williams, C.K., Winn, J.,
  Zisserman, A.: The pascal visual object classes challenge: A retrospective.
  International Journal of Computer Vision  \textbf{111}(1),  98--136 (2015)

\bibitem{guo2020gluoncv}
Guo, J., He, H., He, T., Lausen, L., Li, M., Lin, H., Shi, X., Wang, C., Xie,
  J., Zha, S., et~al.: Gluoncv and gluonnlp: Deep learning in computer vision
  and natural language processing. Journal of Machine Learning Research
  \textbf{21}(23), ~1--7 (2020)

\bibitem{guo2018expandnet}
Guo, S., Alvarez, J.M., Salzmann, M.: {ExpandNet}: Training compact networks by
  linear expansion. arXiv preprint arXiv:1811.10495  (2018)

\bibitem{he2015delving}
He, K., Zhang, X., Ren, S., Sun, J.: Delving deep into rectifiers: Surpassing
  human-level performance on {ImageNet} classification. In: Proceedings of the
  IEEE International Conference on Computer Vision. pp. 1026--1034 (2015)

\bibitem{he2016deep}
He, K., Zhang, X., Ren, S., Sun, J.: Deep residual learning for image
  recognition. In: Proceedings of the IEEE Conference on Computer Vision and
  Pattern Recognition. pp. 770--778 (2016)

\bibitem{he2016identity}
He, K., Zhang, X., Ren, S., Sun, J.: Identity mappings in deep residual
  networks. In: Proceedings of the European Conference on Computer Vision. pp.
  630--645. Springer (2016)

\bibitem{howard2019searching}
Howard, A., Sandler, M., Chu, G., Chen, L.C., Chen, B., Tan, M., Wang, W., Zhu,
  Y., Pang, R., Vasudevan, V., et~al.: Searching for {MobileNetv3}. In:
  Proceedings of the IEEE International Conference on Computer Vision. pp.
  1314--1324 (2019)

\bibitem{howard2017mobilenets}
Howard, A.G., Zhu, M., Chen, B., Kalenichenko, D., Wang, W., Weyand, T.,
  Andreetto, M., Adam, H.: {MobileNets}: Efficient convolutional neural
  networks for mobile vision applications. arXiv preprint arXiv:1704.04861
  (2017)

\bibitem{ioffe2015batch}
Ioffe, S., Szegedy, C.: Batch normalization: Accelerating deep network training
  by reducing internal covariate shift. arXiv preprint arXiv:1502.03167  (2015)

\bibitem{kohler2019exponential}
Kohler, J., Daneshmand, H., Lucchi, A., Hofmann, T., Zhou, M., Neymeyr, K.:
  Exponential convergence rates for batch normalization: The power of
  length-direction decoupling in non-convex optimization. In: The 22nd
  International Conference on Artificial Intelligence and Statistics. pp.
  806--815 (2019)

\bibitem{kohler2018towards}
Kohler, J.M., Daneshmand, H., Lucchi, A., Zhou, M., Neymeyr, K., Hofmann, T.:
  Towards a theoretical understanding of batch normalization. arXiv preprint
  arXiv:1805.10694  (2018)

\bibitem{krizhevsky2009learning}
Krizhevsky, A.: Learning Multiple Layers of Features from Tiny Images. Master's
  thesis, University of Toronto (April 2009)

\bibitem{krizhevsky2012imagenet}
Krizhevsky, A., Sutskever, I., Hinton, G.E.: {ImageNet} classification with
  deep convolutional neural networks. In: Advances in Neural Information
  Processing Systems. pp. 1097--1105 (2012)

\bibitem{li2019selective}
Li, X., Wang, W., Hu, X., Yang, J.: Selective kernel networks. In: Proceedings
  of the IEEE Conference on Computer Vision and Pattern Recognition. pp.
  510--519 (2019)

\bibitem{lin2017focal}
Lin, T.Y., Goyal, P., Girshick, R., He, K., Doll{\'a}r, P.: Focal loss for
  dense object detection. In: Proceedings of the IEEE International Conference
  on Computer Vision. pp. 2980--2988 (2017)

\bibitem{lin2014microsoft}
Lin, T.Y., Maire, M., Belongie, S., Hays, J., Perona, P., Ramanan, D.,
  Doll{\'a}r, P., Zitnick, C.L.: Microsoft coco: Common objects in context. In:
  Proceedings of the European Conference on Computer Vision. pp. 740--755.
  Springer (2014)

\bibitem{Ren2015FasterRT}
Ren, S., He, K., Girshick, R.B., Sun, J.: Faster {R-CNN}: Towards real-time
  object detection with region proposal networks. IEEE Transactions on Pattern
  Analysis and Machine Intelligence  \textbf{39},  1137--1149 (2015)

\bibitem{russakovsky2015imagenet}
Russakovsky, O., Deng, J., Su, H., Krause, J., Satheesh, S., Ma, S., Huang, Z.,
  Karpathy, A., Khosla, A., Bernstein, M., et~al.: {ImageNet} large scale
  visual recognition challenge. International Journal of Computer Vision
  \textbf{115}(3),  211--252 (2015)

\bibitem{salimans2016weight}
Salimans, T., Kingma, D.P.: Weight normalization: A simple reparameterization
  to accelerate training of deep neural networks. In: Advances in Neural
  Information Processing Systems. pp. 901--909 (2016)

\bibitem{sandler2018mobilenetv2}
Sandler, M., Howard, A., Zhu, M., Zhmoginov, A., Chen, L.C.: {MobileNetv2}:
  Inverted residuals and linear bottlenecks. In: Proceedings of the IEEE
  Conference on Computer Vision and Pattern Recognition. pp. 4510--4520 (2018)

\bibitem{santurkar2018does}
Santurkar, S., Tsipras, D., Ilyas, A., Madry, A.: How does batch normalization
  help optimization? In: Advances in Neural Information Processing Systems. pp.
  2483--2493 (2018)

\bibitem{simonyan2014very}
Simonyan, K., Zisserman, A.: Very deep convolutional networks for large-scale
  image recognition. In: ICLR (2014)

\bibitem{szegedy2015going}
Szegedy, C., Liu, W., Jia, Y., Sermanet, P., Reed, S., Anguelov, D., Erhan, D.,
  Vanhoucke, V., Rabinovich, A.: Going deeper with convolutions. In:
  Proceedings of the IEEE Conference on Computer Vision and Pattern
  Recognition. pp.~1--9 (2015)

\bibitem{xie2017aggregated}
Xie, S., Girshick, R., Doll{\'a}r, P., Tu, Z., He, K.: Aggregated residual
  transformations for deep neural networks. In: Proceedings of the IEEE
  Conference on Computer Vision and Pattern Recognition. pp. 1492--1500 (2017)

\bibitem{zhu2019deformable}
Zhu, X., Hu, H., Lin, S., Dai, J.: Deformable convnets v2: More deformable,
  better results. In: Proceedings of the IEEE Conference on Computer Vision and
  Pattern Recognition. pp. 9308--9316 (2019)

\end{thebibliography}

\end{document}